\documentclass[12pt]{article}
\usepackage{amsmath,amssymb,amsfonts}
\usepackage[english]{babel}
\newtheorem{definition}{Definition}[section]
\usepackage{algorithm, algpseudocode}

\usepackage{dsfont}
\usepackage{graphicx}
\usepackage{natbib} 
\usepackage{url} 
\usepackage[english]{babel}

\newcommand{\blind}{0}

\graphicspath{{./figures/}}

\begin{document}

\bibliographystyle{jasa3}

\def\spacingset#1{\renewcommand{\baselinestretch}%
{#1}\small\normalsize} \spacingset{1}


\if0\blind
{
  \title{\bf On tracking varying bounds \\ when forecasting bounded time series \vspace{4mm}}
  \author{Amandine Pierrot$^{\dag,}$\thanks{
    The authors gratefully acknowledge {\O}rsted for providing the data for the Anholt offshore wind farm. The research leading to this work was carried out as a part of the Smart4RES project (European Union’s Horizon 2020, No. 864337). The sole responsibility of this publication lies with the authors. The European Union is not responsible for any use that may be made of the information contained therein.} , Pierre Pinson$^{\sharp,\ddag}$ \vspace{2mm}\\
    $^\dag$ \small Technical University of Denmark, Department of Wind and Energy Systems \vspace{2mm}\\
    $^\sharp$ \small Imperial College London, Dyson School of Design Engineering\\ $^\ddag$ \small Technical University of Denmark, Department of Technology, Management and Economics}
    \date{} 
  \maketitle
} \fi

\if1\blind
{
  \bigskip
  \bigskip
  \bigskip
  \begin{center}
    {\LARGE\bf On tracking varying bounds \\ when forecasting bounded time series}
\end{center}
  \medskip
} \fi

\bigskip
\begin{abstract}
We consider a new framework where a continuous, though bounded, random variable has unobserved bounds that vary over time. In the context of univariate time series, we look at the bounds as parameters of the distribution of the bounded random variable. We introduce an extended log-likelihood estimation and design algorithms to track the bound through online maximum likelihood estimation. Since the resulting optimization problem is not convex, we make use of recent theoretical results on Normalized Gradient Descent (NGD) for quasiconvex optimization, to eventually derive an Online Normalized Gradient Descent algorithm. We illustrate and discuss the workings of our approach based on both simulation studies and a real-world wind power forecasting problem.
\end{abstract}

\noindent%
\vfill

\newpage
\spacingset{2} 
\section{Introduction}
\label{sec:intro}
Many statistical applications involve response variables which are both continuous and bounded. This is especially the case when one has to deal with rates, percentages or proportions, for example when interested in the spread of an epidemic \citep{Guolo2014}, the unemployment rates in a given country \citep{Wallis1987} or the proportion of time spent by animals in a certain activity \citep{Cotgreave1994}. Indeed, proportional data are widely encountered within ecology-related statistical problems, see \citet{Warton2011} among others. Similarly, when forecasting wind power generation, the response variable is also such a continuous bounded variable. Wind power generation is a stochastic process with continuous state space which is bounded from below by zero when there is no wind, and from above by the nominal capacity of the turbine (or wind farm) for high-enough wind speeds. More generally, renewable energy generation from both wind and solar energy are bounded stochastic processes, with the same lower bound (i.e., zero energy production) and different characteristics of their upper bound (since solar energy generation has a time-varying maximum depending on the time of day and time of year), see for example \citet{Pinson2012} and \citet{Bacher2009}.

These continuous bounded random variables call for probability distributions with a bounded support such as the beta distribution, truncated distributions or distributions of transformed normal variables as discussed for example in \citet{Johnson1949}. Very often the response variable is first assumed to lie in the unit interval $(0,1)$ and is then rescaled to any interval $(a,b)$ through the transformation $X=(b-a)\tilde{X}+a$, where $\tilde{X} \in (0,1)$ and $X \in (a,b)$. For applications involving such response variables, these bounds $(a,b)$ are always assumed to be fixed to the same values over the sample or throughout the time series. While this assumption surely makes sense in some cases, we argue it can be misleading and negatively impacts inference when the bounds $(a,b)$ actually vary over time or depending on exogenous variables whilst not being observed. In particular it is highly relevant for energy applications, such as wind power probabilistic forecasting, as in practice the upper bound $b$ may change over time, while being unknown, for example in case of curtailment actions for which information is not available or not reliable. Another application could be the inventory problem of the retailer, see \citet{Laderman1953}. Let $X_t$ be the demand for a certain item at time $t$. Like wind power generation, $X_t$ is double-bounded, from below by zero and from above by the stock available at time $t$, that is by a time-varying upper bound $b_t$. To prepare for demand $X_{t+1}$, the retailer needs to find the quantity they should order in the light of the knowledge they have of the past stocks and demands. Similarly to the problem of forecasting wind power generation, the inventory problem might then involve a double-bounded random variable, the demand for a certain item, which can be regarded as a continuous variable for large quantities being involved, and upper bounded by a bound which may vary over time whilst not being observed, for example in case of supply chain issues, information mismanagement, or just for very large retailers that could not track the evolution of the stocks for each item or could so but would rather benefit from an automatic data-driven tracking.

In both those applications, if the random variable happens to get very close to the upper bound, it might be the case that a higher upper bound would have resulted in a higher wind power generation or item demand. Therefore we do not observe the "true" power generation nor item demand. In that sense one could arguably think of it as being related to censoring and truncation. However we make here a different assumption. While truncation assumes the value of the response variable to be never seen (or recorded) if above the upper bound, and censoring assumes one does not know the exact value but does know it lies above the upper bound, we assume here that an upper bound lower than the "true" response results in squeezing the observed value of the variable, and thus in reshaping the probability distribution of the variable.

There are at least two ways of looking at varying bounds which cannot be observed. One can think of them as latent random variables the distribution of the response variable is conditional on. The main advantage of this approach is its generality and flexibility, with the latent bounds $A$ and $B$ being distributed according to a well-specified probability distribution, which might depend on exogenous variables. Suppose we assume a parametric model with parameter vector $\theta$ for both the bounds and the response variable $X$. Because we do not have access to the realizations $a$ and $b$ of the bounds, the maximization of the likelihood function of the realizations $x$ of $X$ might involve complicated high-dimensional integration, possibly computationally infeasible, and would therefore call for algorithms of the Expectation-Maximization kind \citep{Dempster1977}. Moreover, with such a method and for forecasting applications, one needs to first compute (good enough) forecasts of the bounds in order to be able to forecast the response variable. 

An alternative way of thinking of varying bounds which cannot be observed in the more specific context of time series is to consider them as scaling parameters $a$ and $b$ of the parametric distribution of the bounded response variable, to include them in the parameter vector $\theta$, and to assume the time series to be non-stationary, at least regarding $a$ and $b$. This involves the use of online learning algorithms so that the parameter vector can evolve over time. We will focus in this paper on this setup with an upper varying bound $b$. It would be straightforward to carry the same analysis with a lower varying bound $a$. From now on we will refer to $X$ as $X_t$, as our response variable is now indexed by time $t$. As for the bounded distribution of $X_t$, we use the generalized logit-normal distribution introduced by \citet{Mead1965}. The practical use of any family of distributions depends on the possible variation in its shape, and on the ease with which the distribution can be fitted. The generalized logit-normal distribution is very flexible thanks to three parameters: its location $\mu$, its scale $\sigma^2$ and its shape $\nu$. It relies on a generalization of the logit transform and comes down to the logit-normal distribution when $\nu=1$. 
Because the transformed variable is normally distributed, nice properties can be derived for the original random variable $X_t$. In particular, the probability density function (pdf) of $X_t$ can be expressed as a function of the standard normal density. 

We aim to estimate the full parameter $\theta$ of the pdf of $X_t$ which now includes the upper bound $b$ through Maximum Likelihood Estimation (MLE). The first challenge we need to tackle when dealing with the bound as a parameter in a non-stationary setup is how to handle past observations which are out of the support $(0,b)$ of the bounded distribution of $X_t$ and make the log-likelihood to be infinite. We introduce into the log-likelihood a new term which relies on the sigmoid function to take into account those observations in a "soft" finite way. We choose to call this new log-likelihood the \textit{extended} log-likelihood. The second challenge we need to tackle is that when considering the bound as a parameter, we cannot be in a classical convex optimization setup anymore as the negative log-likelihood appears not to be convex with respect to (w.r.t.) the bound parameter. Therefore we propose to move to the more general quasiconvex optimization setup and use recent results about local quasiconvexity and (Stochastic) Normalized Gradient Descent \citep{Hazan2015} to design a batch algorithm out of Normalized Gradient Descent (NGD) and an online algorithm out of Stochastic Normalized Gradient Descent (SNGD). In addition to these novel quasiconvex algorithms we propose a more classical online convex algorithm which relies on a positive definite approximation of the Hessian. 
We present the statistical parametric model for the time series framework with a varying upper bound $b$ in Section \ref{sec:statmodel} and the corresponding MLE in Section \ref{sec:MLE}. In Section \ref{sec:sim} we perform simulations of synthetic data to run the three algorithms we introduced in Section \ref{sec:MLE}. First we look at their performances when tracking the parameter vector $\theta$ over time, then at their performances when forecasting the probability distribution of the bounded variable. In Section \ref{sec:app} we apply these algorithms to real data in order to provide 10-min-ahead probabilistic forecasts of the wind power generation at Anholt offshore wind farm (Denmark). Finally we discuss the results, the limitations and some prospects of the methodology in Section \ref{sec:discuss}.

\section{Statistical model}
\label{sec:statmodel}
\subsection{Parametric distribution with upper bound $b$ as a parameter}
\label{sec:distribution}
Let $\tilde{X}_t$ be a continuous bounded random variable, $\tilde{X}_t \in (0,1)$, and $X_t$ be the corresponding variable rescaled to $(0,b)$ by applying the transformation $X_t=(b-a)\tilde{X}_t+a=b\tilde{X}_t$ where $a=0$. The generalized logit transform $Y_t \in \mathbb{R}$ of $X_t \in (0,b)$ is given by 
\begin{equation*}
	Y_t=\gamma(X_t/b;\nu)=\log{\frac{(X_t/b)^\nu}{1-(X_t/b)^\nu}}, \quad \nu>0, 
\end{equation*}
where $\nu$ is the shape parameter. When $Y_t$ is distributed according to a Gaussian distribution $\mathcal{N}(\mu,\sigma^2)$, the original variable $X_t/b$ is then distributed according to a generalized logit-normal distribution $L_{\nu}(\mu,\sigma^2)$, see for example \citet{Frederic2008} and \citet{Pinson2012}. By time series we also mean series of dependent observations, therefore we assume the expectation of the normal transform $Y_t$ to be an auto-regressive process of order $p$, that is $\mu_t=\sum_{k=1}^p \lambda_k \gamma(x_{t-k}/b;\nu)$. The pdf of $X_t$ conditional on the previous information set $\mathcal{F}_{t-1}$ (the $\sigma$-algebra generated by $X_1, \dots, X_{t-1}$), with parameter vector $\theta = (\lambda_1,\dots,\lambda_p,\sigma^2,\nu,b)$, is then
\begin{equation}
\label{eq:gln0b}
  p_\theta(x_t \vert \mathcal{F}_{t-1}) =
    \begin{cases}
      \frac{1}{\sqrt{2\pi \sigma^2}}\frac{\nu}{x_t\left(1-(x_t/b)^\nu\right)}\exp\left[-\frac{1}{2}\left(\frac{\gamma(x_t/b;\nu)-\mu_{t}}{\sigma}\right)^2\right] & \text{if $0<x_{t-k}<b$},\\
      0 & \text{otherwise}.
    \end{cases}       
\end{equation}
where $k=0,\dots,p$.

\subsection{Time-dependent log-likelihood function}
We wish to estimate the parameter vector $\theta$ of the pdf $p_\theta(x_t \vert \mathcal{F}_{t-1})$ in \eqref{eq:gln0b} through MLE. In the case of a stationary time series and constant parameter $\theta$, this comes to minimizing the negative log-likelihood objective function 
\begin{equation}
\label{eq:negloglik}
    -l(\theta) = -\sum_{t=p+1}^T \log{p_\theta(x_t \vert \mathcal{F}_{t-1})}
\end{equation}
w.r.t. $\theta$, the data sample $x_1,\dots,x_T$ being fixed, assuming the random variables $X_t$ are i.i.d conditionally on $\mathcal{F}_{t-1}$. 

In a non-stationary setup, estimating the parameter vector $\theta$ comes to estimating a parameter vector $\theta_t$ which varies over time, that is to minimizing a time-dependent negative log-likelihood. For ease of notation and because  the negative log-likelihood is to be minimized w.r.t. $\theta$, let $p_\theta(x_j \vert \mathcal{F}_{j-1})=p_j(\theta)$. The time-dependent negative log-likelihood to be minimized at time $t$ is then
\begin{equation}
\label{eq:negloglik.rect.t}
    -l_t(\theta) = -\frac{1}{t-j_0+1}\sum_{j=j_0}^t  \log{p_j(\theta)}.
\end{equation}
We choose to normalize the time-dependent log-likelihood by the number of observations $t-j_0+1$. This does not change the optimal value obtained when minimizing w.r.t. $\theta$ and leads to more consistent values of the objectives when the number of observations varies. The time-dependent negative log-likelihood in \eqref{eq:negloglik.rect.t} is said to be computed over a moving rectangular window, as all $t-j_0+1$ observations are equally weighted. If we wish to give more weight to the most recent observations we can use instead an exponential forgetting factor $\alpha \in (0,1)$. The time-dependent negative log-likelihood is now said to be computed over a moving exponential window and is 
\begin{equation}
\label{eq:negloglik.ewa.t}
    -l_t(\theta) = -\frac{1}{n_\alpha}\sum_{j=j_0}^t \alpha^{t-j} \log{p_j(\theta)},
\end{equation}
where we use $n_\alpha=\frac{1}{1-\alpha}$ for normalizing the weighted negative log-likelihood. 

\begin{figure}[!ht]
    \centering
    \includegraphics[width=\columnwidth]{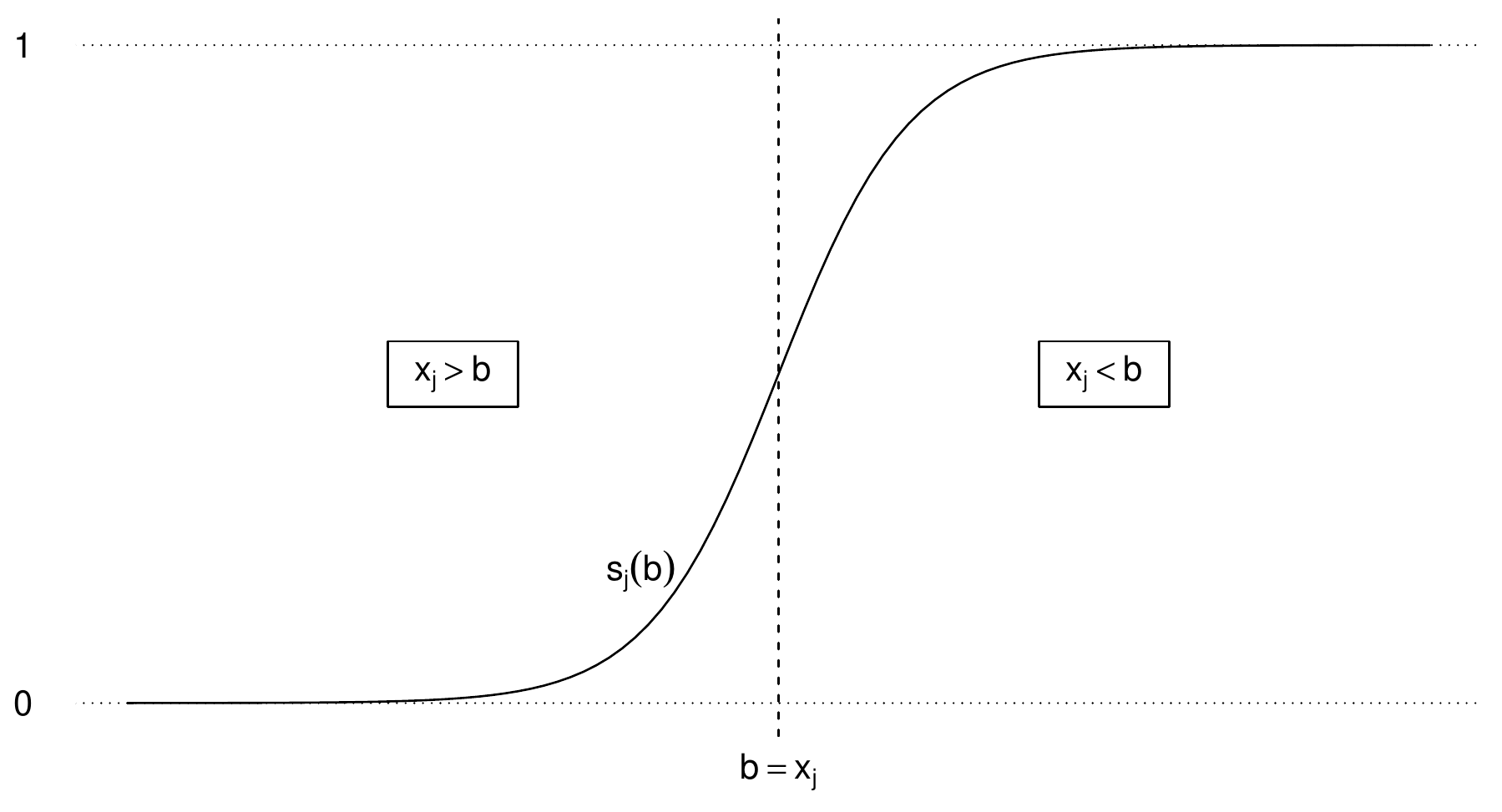}
    \caption{Sigmoid function $s_j(b)$ on the real line.}
    \label{fig:sj}
\end{figure}

From \eqref{eq:gln0b} we can see that the negative log-likelihood in \eqref{eq:negloglik} we wish to minimize takes the value $+\infty$ as soon as an observation $x_t$ is greater or equal to $b$. This is an implicit constraint on $b$ when estimating $\Hat{\theta}$. However moving from the stationary setup to the non-stationary one we do not want $b$ to be greater than all the observations $x_t$ as $b$ should be able to vary over time. Let $U_t=\{j_0, j_0+1,\dots,t\}$, $C_t(\theta)=\{j \in U_t \ \vert \ x_{j-k}<b, k=0,\dots,p \}$ and $\overline{C_t}(\theta)=\{j \in U_t \ | \ j \notin C_t(\theta)\}$ the complement of $C_t(\theta)$ in $U_t$. The log-likelihood takes finite values only for observations $x_j$ such that $j \in C_t(\theta)$. Therefore we can - informally - rewrite  $\displaystyle \sum_{j=j_0}^{t} \alpha^{t-j}\log{p_j(\theta)}$ as $\displaystyle \sum_{j \in C_t(\theta)}^{}\alpha^{t-j}\log{p_j\vert_{b}(\theta)} + \sum_{j \in \overline{C_t}(\theta)}\alpha^{t-j}\log{0}$, where $p_j\vert_{b}(\theta)$ is the pdf $p_j(\theta)$ restricted to its support $(0,b)$, $\alpha=1$ in the case of a rectangular window. When estimating the parameter vector $\theta_t$ over time, we need to take into account all the observations in the past, i.e. even the observations for which the log-likelihood does not take a finite value, that is the observations $x_j$ such that $j$ does not belong to $C_t(\theta)$. We then propose to replace the value 0 in $\log{0}$, which originally corresponds to the value of the pdf $p_j(\theta)$ outside of its support, with a sigmoid function of $b-x_j$
\begin{equation}
    s_j(b)=\frac{1}{1+\exp(-b+x_j)}.
\end{equation}
The function $s_j$ is illustrated in Figure \ref{fig:sj}. 
It can be seen as the probability of $x_j$ to be lower or equal than $b$: we have $s_j(b) \rightarrow 0^+$ when $b << x_j$ and $s_j(b) \rightarrow 1^-$ when $b >> x_j$. Moreover $-\log{s_j(b)}$ is convex and differentiable in $b$. Therefore the "extended" time-dependent negative log-likelihood we propose for a moving rectangular window is 
\begin{equation}
\label{eq:final.negloglik.rect.t}
    -l_t^\infty(\theta) = -\frac{1}{t-j_0+1}\left[\sum_{j \in C_t(\theta)}\log{p_j(\theta)} + \sum_{j \in \overline{C_t}(\theta)}\log{s_j(b)}\right],
\end{equation}
or equivalently for a moving exponential window
\begin{equation}
\label{eq:final.negloglik.ewa.t}
    -l_t^\infty(\theta) = -\frac{1}{n_\alpha}\left[\sum_{j \in C_t(\theta)}\alpha^{t-j} \log{p_j(\theta)} + \sum_{j \in \overline{C_t}(\theta)}\alpha^{t-j}\log{s_j(b)}\right].
\end{equation}
One can note that $j \in \overline{C_t}(\theta)$ does not necessarily mean $x_j \geq b$ as it can happen because a lagged observation $x_{j-k}$ is such that $x_{j-k} \geq b$. In such a case, that is $j \in \overline{C_t}(\theta)$ and $x_j < b$, the observation $x_j$ will still increase the value of the total log-likelihood compared to the event $\{x_j \geq b\}$, which is also a nice feature of choosing this function $s_j$.

\subsection{(Local-)Quasiconvexity}
\label{sec:SLQC}
\begin{figure}[!ht]
    \centering
    \includegraphics[width=\columnwidth]{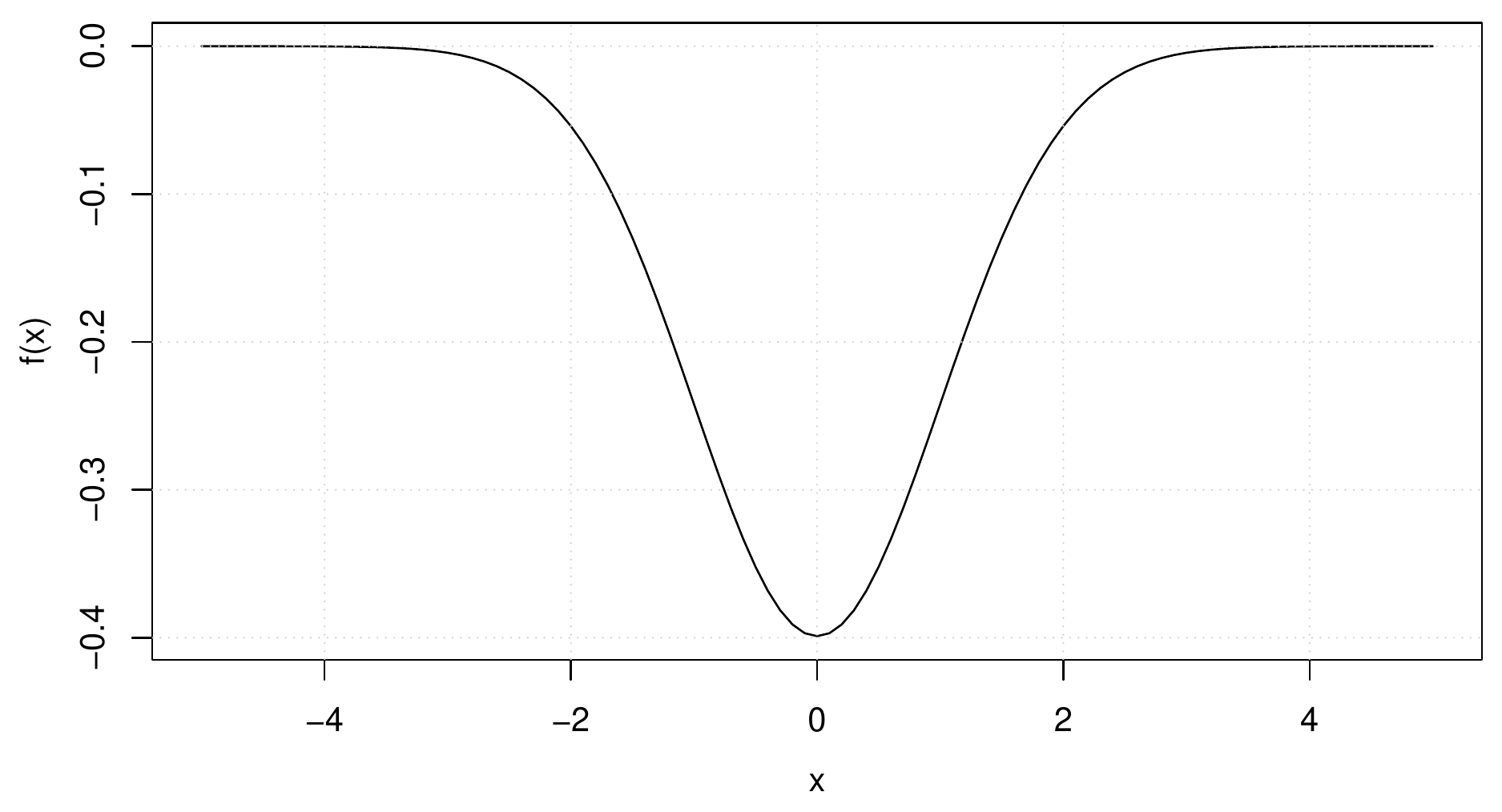}
    \caption{A quasiconvex differentiable function on $\mathbb{R}$, the negative density of a normal variable, with plateau areas when going away from the global minimum.}
    \label{fig:qvcx}
\end{figure}

The first term of the extended time-dependent negative log-likelihoods $-l_t^\infty$ in \eqref{eq:final.negloglik.rect.t} and \eqref{eq:final.negloglik.ewa.t} might not be convex in $\theta$, in particular in $b$. However $-l_t^\infty$ can still have nice properties for it to be globally minimized w.r.t $\theta$. In this section we want to recall a broader class of functions which include convex functions as a subclass: quasiconvex functions. For simplicity let assume functions are differentiable. We use $\|.\|$ to denote the Euclidean norm. From \citet{Boyd2010}, a definition of quasiconvexity is
\begin{definition}[Quasiconvexity]
\label{def:quasicvx1}
A function $f:\mathbb{R}^d \rightarrow \mathbb{R}$ is called quasiconvex (or unimodal) if its domain and all its sublevel sets
\begin{equation*}
    S_\alpha = \{\mathbf{x} \in \mathbf{dom} \ f \ \vert \ f(\mathbf{x})\leq \alpha\},
\end{equation*}
for $\alpha \in \mathbb{R}$, are convex. 
\end{definition}
As an illustrative example, Figure \ref{fig:qvcx} shows the negative probability density function of a normal variable which is a quasiconvex function but not a convex function. 

Quasiconvexity is a considerable generalization of convexity. Still, many of the properties of convex functions hold or have analogs for quasiconvex functions. Following is another definition of quasiconvexity which is equivalent to definition \ref{def:quasicvx1} and an analog to the first-order conditions which hold for convex functions:
\begin{definition}[Quasiconvexity]
\label{def:quasicvx2}
We say that $f:\mathbb{R}^d \rightarrow \mathbb{R}$ is quasiconvex if and only if $\mathbf{dom} \ f$ is convex and for all $\mathbf{x},\mathbf{y} \in \mathbf{dom} \ f$
\begin{equation*}
    f(\mathbf{y}) \leq f(\mathbf{x}) \implies \nabla f(\mathbf{x})^\top(\mathbf{y}-\mathbf{x}) \leq 0.
\end{equation*}
We further say that $f$ is strictly-quasiconvex, if it is quasiconvex and its gradients vanish only at the global minima, i.e. $\displaystyle \forall \ \mathbf{y}: f(\mathbf{y}) > \min_{\mathbf{x} \in \mathbb{R}^d} f(\mathbf{x}) \implies \| \nabla f(\mathbf{y}) \| > 0$.
\end{definition}
However, quasiconvexity broadens but does not fully capture the notion of unimodality in several dimensions. This is the argument of \citet{Hazan2015} who introduce \textit{local-quasiconvexity}, a property that extends quasiconvexity and captures unimodal functions which are not quasiconvex. Let $\mathbb{B}_d(\mathbf{x},r)$ denote the $d$ dimensional Euclidean ball of radius $r$ centered around $\mathbf{x}$, and $\mathbb{B}_d:=\mathbb{B}_d(0,1)$. The definition of local-quasiconvexity as introduced by \citet{Hazan2015} is the following:
\begin{definition}[Local-quasiconvexity]
\label{def:local-quasicvx}
Let $\mathbf{x},\mathbf{z} \in \mathbb{R}^d$, $\kappa,\epsilon >0$. \newline We say that $f:\mathbb{R}^d \mapsto \mathbb{R}$ is $(\epsilon,\kappa,\mathbf{z})$-Strictly-Locally-QuasiConvex (SLQC) in $\mathbf{x}$, if at least one of the following applies:
\begin{enumerate}
    \item $f(\mathbf{x})-f(\mathbf{z}) \leq \epsilon$.
    \item $\| \nabla f(\mathbf{x}) \| >0$, and for every $\mathbf{y} \in \mathbb{B}_d(\mathbf{z},\epsilon / \kappa)$ it holds that $\nabla f(\mathbf{x})^\top(\mathbf{y}-\mathbf{x}) \leq 0$.
\end{enumerate}
\end{definition}
When considering the Generalized Linear Models (GLM) regression as a fitting problem in which we want to minimize the error function 
\begin{equation}
\label{eq:idealGLM}
    \widehat{\text{err}}_m(\mathbf{w})=\frac{1}{m}\sum_{i=1}^m(y_i-\phi\langle\mathbf{w},\mathbf{x}_i\rangle)^2,
\end{equation}
where $(\mathbf{x}_i,y_i)_{i=1,\dots,m}\in \mathbb{B}_d \times [0,1]$ and $\phi: \mathbb{R}\mapsto \mathbb{R}$ is an activation function, \citet{Hazan2015} show that if $\phi$ is the sigmoid function and if we are guaranteed to have $\mathbf{w}^* \in \mathbb{R}^d$ such that $y_i=\phi\langle \mathbf{w}^*,\mathbf{x}_i \rangle$, $\forall i=1,\dots,m$, then the error function in \eqref{eq:idealGLM} is not generally quasiconvex but is indeed SLQC. This setup is said to be the idealized GLM setup. In the more common noisy GLM setup \citep{McCullagh1989}, where we assume now $(\mathbf{x}_i,y_i)_{i=1,\dots,m}$ are i.i.d. samples from an unknown distribution $\mathcal{D}$ and there exists a predictor $\mathbf{w}^*$ such that $\mathbb{E}[y \vert \mathbf{x}]=\phi\langle \mathbf{w}^*,\mathbf{x}\rangle$, $\mathbf{w}^*$ can be shown to be a global minima of the expected error 
\begin{equation}
\label{eq:noisyGLM}
    \mathcal{E}(\mathbf{w})=\mathbb{E}(y-\phi\langle\mathbf{w},\mathbf{x}\rangle)^2.
\end{equation}
Given $m$ samples from $\mathcal{D}$, their empirical error $\widehat{\text{err}}_m(\mathbf{w})$ is defined as in \eqref{eq:idealGLM} and \citet{Hazan2015} show that it is also SLQC, with high probability.

Simulations of $-l_t^\infty$ in \eqref{eq:final.negloglik.rect.t} and \eqref{eq:final.negloglik.ewa.t} show our extended negative log-likelihood not to be convex but rather be, with high probability, a quasiconvex function with plateau areas when $b$ is away from the optimal value $b^*$ and steep concave cliffs in the neighborhood of $b^*$. Therefore it seems reasonable to assume $-l_t^\infty$ to be quasiconvex in $\theta$, or at least locally-quasiconvex.

\section{Time-dependent maximum likelihood estimation}
\label{sec:MLE}
\subsection{Normalized Gradient Descent}
\label{sec:MLE.NGD}
Let $f$ be the quasiconvex objective function we wish to minimize w.r.t parameter $\textbf{x} \in \mathbb{R}^d$. It is well known that quasiconvex problems can be solved through a series of convex feasibility problems \citep{Boyd2010}. However solving such feasibility problems can be very costly and involves finding a family of convex functions $\phi_t:\mathbb{R}^{d} \rightarrow \mathbb{R}$, $t \in \mathbb{R}$, that satisfy 
\begin{equation*}
    f(\mathbf{x})\leq t \iff \phi_t(\mathbf{x}) \leq 0,
\end{equation*}
and $\phi_s(\mathbf{x}) \leq \phi_t(\mathbf{x})$ whenever $s \geq t$, which is far from straightforward in our case, i.e. with $f(\mathbf{x}):=-l_t^\infty(\theta)$. In the batch setup, a pioneering paper by \citet{Nesterov1984} was the first to propose an efficient algorithm, the Normalized Gradient Descent (NGD), and to prove that this algorithm converges to an $\epsilon$-optimal solution within $O(1/\epsilon^2)$ iterations given a differentiable quasiconvex objective function. Gradient descent with fixed step sizes is known to perform poorly when the gradients are too small in a plateau area of the function or explode in cliff areas. Among the deep learning community, there have been several attempts to tackle plateaus and cliffs. However those works do not provide a theoretical analysis showing better convergence guarantees than NGD. 

NGD is presented in Algorithm \ref{algo:NGD}.
\begin{algorithm}[!ht]
\caption{Normalized Gradient Descent (NGD)}\label{algo:NGD}
\begin{algorithmic}
\Require \#Iterations $I, \mathbf{x_1} \in \mathbb{R}^d$, learning rate $\eta$
\For{$i=1,\dots,I$}
\State Update: $\mathbf{x}_{i+1}=\mathbf{x}_{i} - \eta \Hat{g}_i$ where $g_i=\nabla f(\mathbf{x}_i)$, $\Hat{g}_i = \frac{g_i}{\| g_i \|}$
\EndFor
\Ensure $\displaystyle \overline{\mathbf{x}}_I = \arg \min_{\mathbf{x}_1,\dots, \mathbf{x}_I} f(\mathbf{x}_i)$
\end{algorithmic}
\end{algorithm}
It is similar to Gradient Descent, except one normalizes the gradient. It is intuitively clear that to achieve robustness to plateaus (with vanishing gradients) and cliffs (with exploding gradients), one must ignore the size of the gradient. It is more surprising that the information in the direction of the gradient is enough to guarantee convergence. Having introduced SLQC functions, \citet{Hazan2015} prove that NGD also finds an $\epsilon$-optimal minimum for such functions in $O(1/\epsilon^2)$ iterations. They even show faster convergence rates for quasiconvex objective functions which are locally-smooth.

The adaptation of Algorithm \ref{algo:NGD} to our setup is straightforward taking $f(\textbf{x}):=-l_t^\infty(\theta)$ and $I:=t$. However one can note that when minimizing $-l_t^\infty$ w.r.t $\theta=(\Lambda,\sigma^2,\nu,b)$ we shall recover positive estimates of the scale parameter $\sigma^2$ and the shape parameter $\nu$. This is constrained optimization which can be easily overcome by a change of variable such as replacing $\sigma^2$ with $\omega=\log{\sigma^2}$ and $\nu$ with $\tau=\log{\nu}$. 

\subsection{Recursive maximum likelihood estimation}
\label{sec:rMLE}
In the above, we presented the NGD algorithm in order to find a global minimum to $-l_t^\infty$ w.r.t $\theta$ in a batch setting. This implies to run NGD every time we want to update the parameters $\theta$ to account for new observations which is computationally prohibitive. In order to move from the batch setting of the NGD algorithm to the online learning setting we first consider a (classic) recursive MLE procedure to recover time-dependent estimates $\Hat{\theta}_t$ through the time-dependent negative log-likelihood $-l_t^\infty$. It is worth noting that this recursive procedure relies on a Newton step whereas our function $-l_t^\infty$ is with high probability not convex. However it is more a quasi-Newton approach as it approximates the Hessian with a positive definite matrix thanks to first-order information, i.e. gradients. Because we are in the online learning setting we can hope for this approximate to be a good enough very local approximation. Before introducing the recursive MLE procedure let finally note that the algorithm we derive from the latter is related to a quasi-Newton algorithm from online convex optimization (OCO), namely Online Newton Step (ONS), see for example \citet{Hazan2022}. 

Consider the extended time-dependent log-likelihood with a moving exponential window in \eqref{eq:final.negloglik.ewa.t} and recall that $n_\alpha = \frac{1}{1-\alpha}$. We can rewrite \eqref{eq:final.negloglik.ewa.t} as
\begin{equation}
\label{eq:rec}
    -l_t^\infty(\theta)=
    \begin{cases}
        -\alpha l_{t-1}^\infty(\theta)-(1-\alpha)\log{p_t^{}(\theta)} & \text{if $t \in C_t(\theta)$},\\
      -\alpha l_{t-1}^\infty(\theta)-(1-\alpha)\log{s_t^{}(b)} & \text{if $t \in 
      \overline{C_t}(\theta)$}.
    \end{cases}
\end{equation}
Let now $\Hat{\theta}_{t}$ be the estimate of the parameter vector at time $t$. The recursive MLE procedure relies on a Newton step for obtaining the estimate $\Hat{\theta}_t$ as a function of the previous estimate $\Hat{\theta}_{t-1}$, see for example \citet{Madsen2007} and \citet{Madsen2012}. Applying one Newton step at time $t$ we have
\begin{equation}
\label{eq:step}
    \Hat{\theta}_t = \Hat{\theta}_{t-1} - \frac{\nabla_\theta^{} l_t^\infty(\Hat{\theta}_{t-1})}{\nabla^2_\theta l_t^\infty(\Hat{\theta}_{t-1})}.
\end{equation}
Let $\textbf{h}_{t} = \nabla_\theta \log{p_t(\Hat{\theta}_{t-1})}$ if $t \in C_t(\Hat{\theta}_{t-1})$, $\mathbf{h}_{t} = \nabla_\theta \log{s_t(\Hat{b}_{t-1})}$ if $t \in \overline{C_t}(\Hat{\theta}_{t-1})$ and $\Hat{\mathbf{R}}_t = -\nabla^2_\theta l_t^\infty(\Hat{\theta}_t)$. The recursive estimation relies on a few additional assumptions which are classic in the online learning framework.  First assuming $\Hat{\theta}_{t-1}$ minimizes $-l_{t-1}^\infty(\theta)$, from \eqref{eq:rec} we get 
\begin{equation}
\label{eq:gradient}
    \nabla_\theta^{} l_t^\infty(\Hat{\theta}_{t-1})= (1-\alpha)\mathbf{h}_t.
\end{equation}
Then assuming $p_t$ and $s_t$ are (almost) linear in $\Gamma$ in the neighborhood of $\Hat{\Gamma}_{t-1}$, we get the approximation $\nabla^2_\theta \log{p_t(\Hat{\theta}_{t-1})} = -\mathbf{h}^{}_{t} \mathbf{h}^\top_{t}$ if $t \in C_t(\Hat{\theta}_{t-1})$ or $\nabla^2_\theta \log{s_t(\Hat{b}_{t-1})} = -\mathbf{h}^{}_{t} \mathbf{h}^\top_{t}$ if $t \in \overline{C_t}(\Hat{\theta}_{t-1})$. These approximations, which can be made because $\log{p_t}$ and $\log{s_t}$ are logarithms, are the key for ensuring that the approximate $\Hat{\mathbf{R}}_t$ of the Hessian matrix is always positive definite. Finally we assume that the objective criterion $-l_t^\infty$ is smooth in the vicinity of $\Hat{\theta}_{t}$, and the adaptation step small enough so that $ \Hat{\mathbf{R}}_t =-\nabla^2_\theta l_t^\infty(\Hat{\theta}_t) \simeq -\nabla^2_\theta l_t^\infty(\Hat{\theta}_{t-1})$. This is a classic assumption for deriving recursive estimation methods for stochastic systems \citep{Ljung1983}. Our two-step recursive scheme at time $t$ is then 
\begin{align*}
    \Hat{\mathbf{R}}_t &= \alpha \Hat{\mathbf{R}}_{t-1} + (1-\alpha)\mathbf{h}^{}_{t} \mathbf{h}^\top_{t}, \\
    \Hat{\theta}_t &= \Hat{\theta}_{t-1} + (1-\alpha) \Hat{\mathbf{R}}_t^{-1}\mathbf{h}^{}_{t}. 
\end{align*}

An algorithm based upon such a recursive scheme might face computational issues as it requires inverting a matrix, the information matrix  $\Hat{\mathbf{R}}_t$, at each iteration. This can be prevented by working directly with the matrix inverse, the covariance matrix $\Hat{\mathbf{P}}_t$, which can be computed by using the matrix inversion rule. The detailed computations are available in the supplementary material and the resulting algorithm is described in Algorithm \ref{algo:rMLE.b}.
\begin{algorithm}[!ht]
\caption{Recursive Maximum Likelihood Estimation (rMLE)}\label{algo:rMLE.b}
\begin{algorithmic}
\Require $T, \theta_{p} \in \mathbb{R}^{p+3}$, forgetting factor $\alpha \in (0,1)$, $\Hat{\mathbf{P}}_{p}=10^6\mathbf{I}_{p+3}$
\For{$t=p+1,\dots,T$}
\State Set  $\mathbf{h}^{}_{t} = \nabla_\theta \log{p_t(\Hat{\theta}_{t-1})}$ if $t \in C_t(\Hat{\theta}_{t-1})$ or set $\mathbf{h}^{}_{t} = \nabla_\theta \log{s_t(\Hat{\theta}_{t-1})}$ if $t \in \overline{C_t}(\Hat{\theta}_{t-1})$.
\State Update: 
\begin{align*}
    \Hat{\mathbf{P}}_t &= \frac{1}{\alpha}\left[\mathbf{I}_{p+3} - \frac{\Hat{\mathbf{P}}_{t-1}^{}\mathbf{h}^{}_{t} \mathbf{h}^\top_{t}}{\frac{\alpha}{1-\alpha}+\mathbf{h}^\top_{t}\Hat{\mathbf{P}}_{t-1}^{}\textbf{h}^{}_{t}}\right] \Hat{\mathbf{P}}_{t-1} \\
    \Hat{\theta}_t^{} &= \Hat{\theta}_{t-1}^{} + (1-\alpha) \Hat{\mathbf{P}}_t^{}\mathbf{h}_{t}^{}
\end{align*}
\EndFor
\end{algorithmic}
\end{algorithm}

\subsection{Online Normalized Gradient Descent}
\label{sec:MLE.ONGD}
The recursive MLE algorithm described in the former section is an OCO algorithm. While it can be used for solving our quasiconvex optimization problem in an online learning setting it is likely that it would do it in a suboptimal way. Therefore we propose to alternatively use Online Normalized Gradient Descent (ONGD) as an Online QuasiConvex Optimization (OQCO) algorithm. From the observation that online learning and stochastic optimization are closely related and interchangeable (see for example \citet{CesaBianchi2004} and \citet{Duchi2011}), we use the Stochastic Normalized Gradient Descent (SNGD) introduced by \citet{Hazan2015} to deriving the corresponding Online Normalized Gradient Descent for online learning. 

Recall the definition \ref{def:local-quasicvx} of local-quasiconvexity in section \ref{sec:SLQC}. Taking advantage of the SLQC assumption, \citet{Hazan2015} present SNGD, which is similar to Stochastic Gradient Descent (SGD) except they normalize the gradients, and prove the convergence of SNGD within $O(1/\epsilon^2)$ iterations to an $\epsilon$-optimal minimum. This positive result requires that at each iteration of SNGD the gradient should be estimated using a minibatch of minimal size $m$. Indeed the authors provide a negative result showing that if the minibatch size is too small then the algorithm might diverge. This is where SNGD differs again from SGD as in the latter and for the case of convex functions even a minibatch of size 1 is enough for guaranteed convergence. The general ONGD derived from SNGD is presented in Algorithm \ref{algo:ONGD}. To the best of our knowledge, this is the first time SNGD is used in an online learning fashion for OQCO. 

\begin{algorithm}[!ht]
\caption{Online Normalized Gradient Descent (ONGD)}\label{algo:ONGD}
\begin{algorithmic}
\Require convex set $\mathcal{K}, T, \mathbf{x}_m \in \mathcal{K}$, step size $\eta$, minibatch size $m$
\For{$t=m,\dots,T$}
\State Play $\mathbf{x}_{t}$ and observe cost $\displaystyle f_t(\mathbf{x}_{t})=\frac{1}{m}\sum_{j=t-m+1}^{t}f_{j}(\mathbf{x}_{t})$.
\State Update and project:
\begin{align*}
    \mathbf{y}_{t+1} &= \mathbf{x}_{t} - \eta \Hat{g}_t \ \text{where} \ g_t=\nabla f_t(\mathbf{x}_t), \ \Hat{g}_t = \frac{g_t}{\| g_t \|}\\
    \mathbf{x}_{t+1} &= \Pi_\mathcal{K}(\mathbf{y}_{t+1})
\end{align*}
\EndFor
\end{algorithmic}
\end{algorithm}

Following the framework introduced by \citet{CesaBianchi2006}, ONGD is defined in terms of a repeated game played between the online player and the "environment" generating the outcome sequence. At each iteration $t$, we play a parameter vector $\mathbf{x}_t$. After we have committed to this choice, a (SLQC) cost function $f_t$ is revealed and the cost we incur is therefore $f_t(\mathbf{x}_t)$, the value of the cost function for the choice $\mathbf{x}_t$. Similarly to Online Gradient Descent (OGD), which is based on standard gradient descent from offline optimization and was introduced in its online form by \citet{Zinkevich2003}, we have included in ONGD a projection step $\Pi_\mathcal{K}(.)$. Indeed in each iteration, the algorithm takes a step from the previous point in the direction of the normalized gradient of the previous cost. This step may result in a point outside of the underlying convex set $\mathcal{K}$. In such cases the algorithm therefore projects the point back to the convex set $\mathcal{K}$, i.e. finds its closest point in $\mathcal{K}$.

The adaptation of Algorithm \ref{algo:ONGD} to our setup is straightforward taking
\begin{equation*}
    f_{j}(\mathbf{x}_t):=
    \begin{cases}
        -\log{p_j(\Hat{\theta}_t)}  & \text{if $j \in C_t(\Hat{\theta}_{t})$},\\
        -\log{s_j(\Hat{b}_t)} & \text{if $j \in \overline{C_t}(\Hat{\theta}_{t})$}.
    \end{cases}
\end{equation*}
Note that working with $\theta=(\Lambda,\omega,\tau,b)$, $\theta \in \mathbb{R}^{p+3}$, as in sections \ref{sec:MLE.NGD} and \ref{sec:rMLE}, we do not need the projection step in our setup. Finally we want to emphasize that moving from Algorithms \ref{algo:NGD} and \ref{algo:rMLE.b} to Algorithm \ref{algo:ONGD} we do not need a moving window anymore. Instead we use a constant step size $\eta$, analog to the learning rate in NGD and SNGD, and a minibatch size $m$. The technical derivations required for all the algorithms of section \ref{sec:MLE} are available in the supplementary material.

\section{Simulation study}
\label{sec:sim}
\subsection{Tracking the parameter vector}
\label{sec:sim.tracking}
In order to test the convergence of the algorithms proposed in section \ref{sec:MLE} we perform an empirical study on synthetic data which fit our setup. We run 100 Monte Carlo (MC) simulations with $T=12000$, $\lambda=0.9$, $\sigma^2=1$, $\nu=1.5$ and $b$ varying in a sinusoidal way. For all algorithms and simulations the lag $p=1$ of the auto-regressive process is assumed to be known and the initial values of the parameter vector $\theta=(\lambda,\sigma^2,\nu,b)$ are always set to $(0,1,1,1)$. For NGD each batch algorithm is run every 500 data points after a burn-in period of 1,000 points. A new estimate of the parameter vector is therefore available every 500 time steps. For rMLE in order to fulfill the necessary condition to \eqref{eq:gradient} in section \ref{sec:rMLE}, the recursive algorithm is run after a warm-up period of 1,000 data points: first a batch algorithm, i.e. NGD, is run on the first 1,000 data points with $\theta_0=(0,1,1,1)$; then the resulting estimates are used as initial values for rMLE. From the 1,000th time step on, a new estimate of the parameter vector is then available every time step. For ONGD the algorithm starts as soon as there are enough data points for a minibatch of size $m$. Afterwards a new estimate of the parameter vector is available every time step. The values of the hyperparameters for each algorithm are summarized in Table \ref{tab:hyperpar}. They were decided upon looking to the first MC simulation. 
\begin{table}[!ht]
\caption{Hyperparameter values for each algorithm: forgetting factor $\alpha$, number of iterations $I$, learning rate/step size $\eta$ and minibatch size $m$.}
\begin{center}
\begin{tabular}{lcccc}
& $\alpha$ & $I$ & $\eta$ & $m$\\
\hline
NGD (Algorithm \ref{algo:NGD}) &0.990 &10000 &0.003 &- \\
rMLE (Algorithm \ref{algo:rMLE.b}) &0.975&- &- &- \\
ONGD (Algorithm \ref{algo:ONGD}) &- &- &0.001 &100 \\
\end{tabular}
\label{tab:hyperpar}
\end{center}
\end{table}

\begin{figure}[!ht]
    \centering
    \includegraphics[width=\columnwidth]{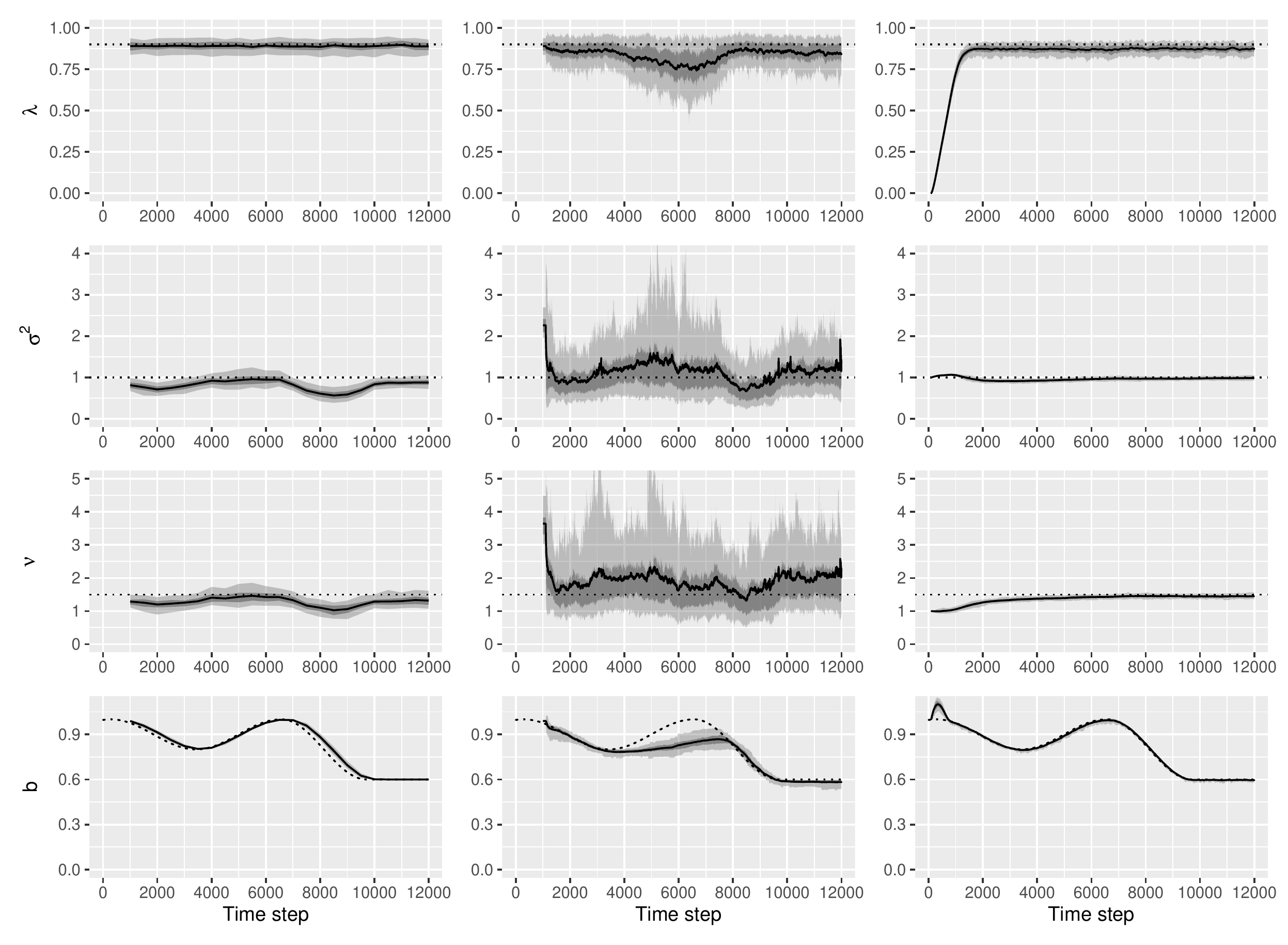}
    \caption{Confidence intervals of the tracked parameters for NGD (left), rMLE (center) and ONGD (right) with coverage probabilities 0.9 and 0.5, along with the average estimates (solid lines) and the true parameters (dotted lines).}
    \label{fig:tracking}
\end{figure}

The tracking of the parameters depending on the algorithm is presented in Figure \ref{fig:tracking}. First note that while all three algorithms succeed in tracking the parameter vector, there are clear differences in their performances. ONGD is the algorithm which manages to closer track the parameters with the less variance. However for some parameters it needs a few time steps before converging to the true value at the time: about 1,000 for the auto-regressive parameter $\lambda$, 2,000 for the shape parameter $\nu$ and 500 for the upper bound $b$. The scale parameter $\sigma^2$ is already at its true value when the algorithms get started but serves as a reference that there is no divergence from the optimal value. As a matter of fact and as stated by \citet{Hazan2015} we have observed the parameter $\sigma^2$ to diverge when the minibatch size $m$ was too small: for $m=1$ the algorithm was surely diverging while it was converging for $m \geq 10$. NGD converges from the first update already, that is with 1,000 data points. This was to be expected as $\alpha$ was set to 0.99, which means most of the weight was placed on the last 100 points of each batch. However NGD shows more variance than ONGD and is a bit slower in following a decreasing $b$. The latter drawback would likely be improved by increasing the frequence in the updates. Finally rMLE performs well in tracking a decreasing $b$, but fails to properly track an increasing bound. Moreover this comes with a cost in variance which is very high.

\subsection{Forecasting the distribution}
\label{sec:sim.forecasting}
Because many applications which might benefit from this framework involve forecasting, we are now interested in the performance of the algorithms when forecasting at time $t$ the distribution of the bounded variable $X_{t + 1}$. To be able to track the bound parameter over time, we have introduced in section \ref{sec:statmodel} the extended time-dependent negative log-likelihood $-l_t^\infty$ and allowed $b_t$ to vary on $\mathds{R}$, which makes sense from an inference point of view. Because we are working with series of dependent observations, when moving to forecasting the distribution of $X_{t+1}$ we need the current value $b_t$ to be greater than all the $p$ former observed values of $X_t,\dots,X_{t-p+1}$ for the expected value of $X_{t+1}$ to exist. Therefore we introduce a projection step as described in section \ref{algo:ONGD} for ONGD: we project $\Hat{\theta}_t$ on the convex set $\mathcal{K}=\mathds{R}^{p+2} \times (\max (x_{t},\dots,x_{t-p+1}),+\infty)$ and we get the projected parameter $\Tilde{\theta}_t = \Pi_\mathcal{K}(\Hat{\theta}_t)=(\Hat{\Lambda}_t,\Hat{\omega}_t,\Hat{\tau}_t,\Tilde{b}_t)$ where $\Tilde{b}_t=\max (x_{t},\dots,x_{t-p+1})+\delta$ if $\max (x_{t},\dots,x_{t-p+1}) > \Hat{b}_t$, $\Tilde{b}_t=\Hat{b}_t$ if $\max (x_{t},\dots,x_{t-p+1}) < \Hat{b}_t$. Note that we need to introduce a small $\delta > 0$ as we project $\Hat{b}_t$ on an open convex set. When looking at the observation $x_t$ as a coarsened version of $X_t$, $\delta$ can be seen as a coarsening parameter. This coarsened data framework has been formalized by \citet{Heitjan1991} and \citet{Heitjan1993}. We use $\delta=0.001$.

The predictive probability distributions obtained from Algorithms \ref{algo:NGD}, \ref{algo:rMLE.b}, \ref{algo:ONGD} are evaluated and compared to classic benchmarks such as climatology and probabilistic persistence. The climatology is based on all past data available at the time of forecasting and the probabilistic persistence is the most recent observed value at the time of forecasting which we dress with the most recent observed values of the persistence error. We also provide the predictive distributions obtained from the alike algorithm to Algorithm \ref{algo:rMLE.b} when the bound is assumed to be fixed and equal to 1 \citep{Pierrot2021}. From now on we will refer to this algorithm as rMLE.1 and to Algorithm \ref{algo:rMLE.b} as rMLE.b. We evaluate the predictive distributions through calibration and proper scoring rules, as probabilistic predictions should be calibrated and as informative/sharp as possible, see for example \citet{Gneiting2007a}, \citet{Gneiting2007b} and \citet{Gneiting2014}. To empirically check on probabilistic, respectively marginal calibration, we provide Probability Integral Transform (PIT) histograms, respectively marginal calibration plots. Scoring rules are attractive measures of predictive performance as they evaluate calibration and sharpness simultaneously. We use the Continuous Ranked Probability Score (CRPS) which is a proper scoring rule relative to the class $\mathcal{P}$ of the Borel probability measures on $\mathds{R}$ and a strictly proper scoring rule relative to the subclass $\mathcal{P}_1$ of the Borel probability measures that have finite first moment \citep{Gneiting2007b}. Proper scoring rules are often used in negative orientation, e.g. the lower the better. As we work with predictive densities, one could think of using the logarithmic score which is strictly proper relative to all measures that are absolutely continuous, at least to compare the predictive densities provided by Algorithms \ref{algo:NGD}, \ref{algo:rMLE.b}, \ref{algo:ONGD} and the rMLE.1 benchmark algorithm. However in our framework it can always happen that $x_{t+1}$ falls out of the support of the predictive density $\Hat{p}_{t+1}$ we have produced at time $t$, when the current estimate of the upper bound is too low. In such a case the logarithmic score is equal to $-\log{\Hat{p}_{t+1}(x_{t+1})}=-\log 0 = +\infty$, which is not suitable. In contrast the CRPS is defined on $\mathds{R}$ as
\begin{equation}
\label{eq:CRPS}
    \text{CRPS}(F,x) = \int_{-\infty}^{\infty}\left(F(y)-\mathds{1}_{y \geq x}\right)^2\text{d}y,
\end{equation}
where $F$ is the cumulative distribution function (cdf) of the probabilistic forecast and y is the evaluation point. In our setup $F:=\Hat{F}_{t+1}$ and $x:=x_{t+1}$. If $x_{t+1}$ happens to be greater than the upper bound $b_t$ we get 
\begin{align*}
    \text{CRPS}(\Hat{F}_{t+1},x_{t+1}) &= \int_{-\infty}^{b_t}\left(\Hat{F}_{t+1}(y)-\mathds{1}_{y \geq x_{t+1}}\right)^2\text{d}y + \int_{b_t}^{\infty}\left(1-\mathds{1}_{y \geq x_{t+1}}\right)^2\text{d}y, \\
    &= \int_{-\infty}^{b_t}\Hat{F}_{t+1}(y)^2\text{d}y + \int_{b_t}^{x_{t+1}}1 \ \text{d}y + \int_{x_{t+1}}^\infty 0 \ \text{d}y, \\
    &= \int_{-\infty}^{b_t}\Hat{F}_{t+1}(y)^2\text{d}y + x_{t+1} - b_t.
\end{align*}
Therefore the CRPS is increased by an observation falling out of the support of the predictive distribution but to a higher finite value contrary to the logarithmic score which becomes infinite. Moreover the CRPS allows us to compare discrete and continuous distributions, that is to compare our density-based algorithms to climatology and probabilistic persistence, as if the predictive distribution takes the form of a sample of size $N$, then the right side of \eqref{eq:CRPS} can be evaluated in $\mathcal{O}(N\log{N})$ operations \citep{Hersbach2000}.

\begin{figure}[!ht]
    \centering
    \includegraphics[width=\columnwidth]{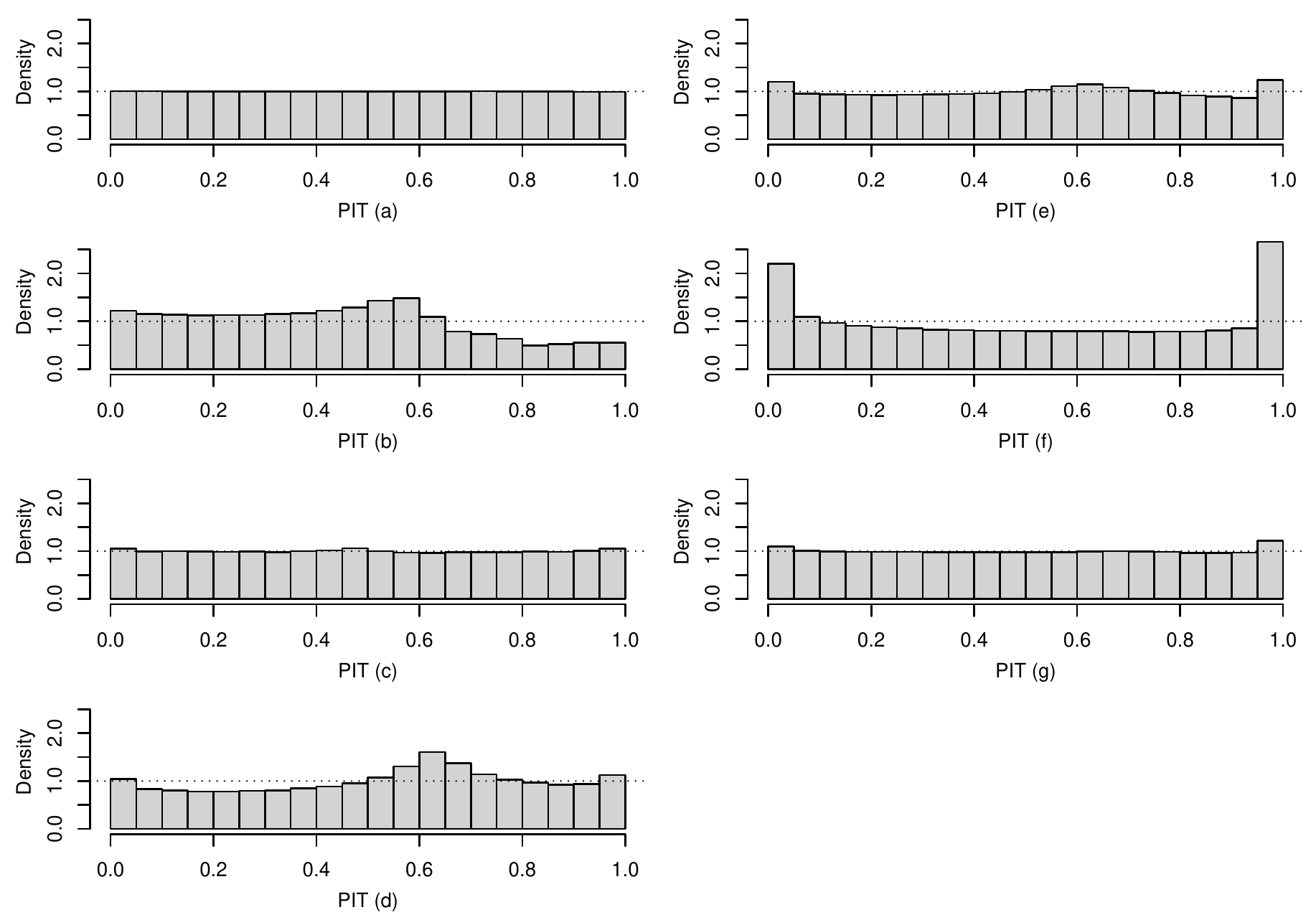}
    \caption{PIT histograms for all benchmarks (left) and Algorithms \ref{algo:NGD}, \ref{algo:rMLE.b}, \ref{algo:ONGD} (right): (a) ideal forecaster, (b) climatology, (c) probabilistic persistence, (d) rMLE.1, (e) NGD, (f) rMLE.b, (g) ONGD.}
    \label{fig:PIT}
\end{figure}

\begin{figure}[!ht]
    \centering
    \includegraphics[width=\columnwidth]{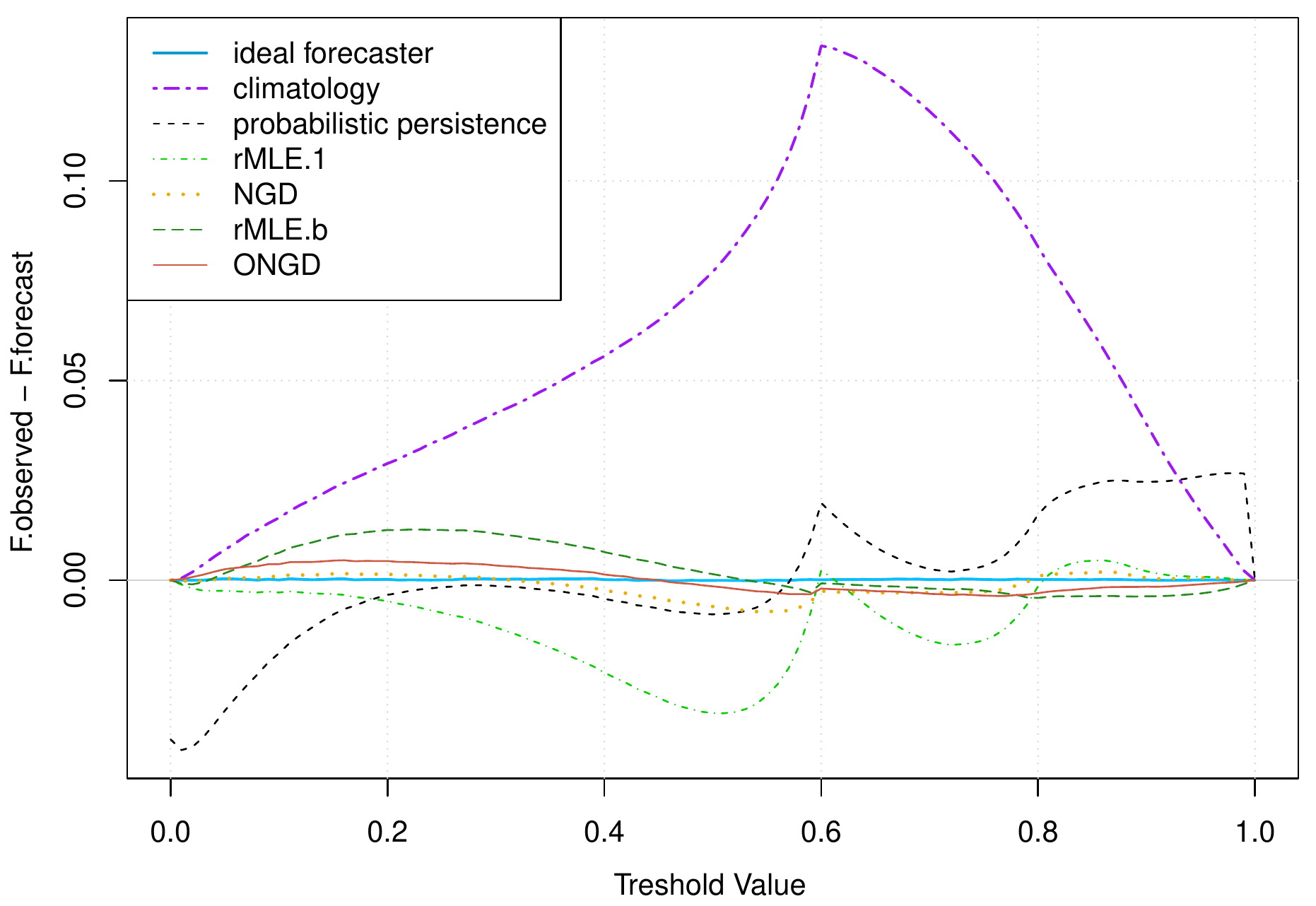}
    \caption{Marginal calibration plot for all benchmarks and Algorithms \ref{algo:NGD}, \ref{algo:rMLE.b}, \ref{algo:ONGD}.}
    \label{fig:marginal}
\end{figure}

We start computing predictive distributions after having seen 2,000 observations. Recall that the hyperparameters were chosen in section \ref{sec:sim.tracking} upon looking at only the first MC simulation for each algorithm, but the whole simulated time series, that is looking at the data we are now computing probabilistic forecasts for. This may be optimistic, even if we only looked at the data from the first MC simulation. In order for our algorithms to not be more optimistic than the benchmarks, we then use the hyperparameters for probabilistic persistence and rMLE.1  that gave the best CRPS on the first MC simulation. We also provide the results for the ideal forecaster, which is the true distribution of the synthetic data, that is the GLN distribution with the true constant values $\Lambda$, $\sigma^2$ and $\nu$ and the true values of the upper bound $b_{t+1},\dots,b_{t-p+1}$. The CRPS.s are available in Table \ref{tab:CRPS}, PIT histograms with 20 bins in Figure \ref{fig:PIT}, and the marginal calibration plot in Figure \ref{fig:marginal}. Note that they are all averages on the MC sample.

\begin{table}[!ht]
\caption{1-step-ahead CRPS and respective improvement over climatology and persistence. The CRPS is averaged over the MC sample, and the standard deviation is also provided.}
\begin{center}
\begin{tabular}{lrrr}
 &mean (sd) &Imp./clim. &Imp./persist.\\
\hline
ideal forecaster &5.78\% (0.10) &- &-\\
\hline
climatology &15.26\% (0.24) &- &-\\
probabilistic persistence &6.28\% (0.10) &58.85\% &- \\
rMLE.1 &6.04\% (0.09) &60.40\% &3.77\%\\
\hline
NGD (Algorithm \ref{algo:NGD}) &5.87\% (0.09) &61.53\% &6.52\%\\
rMLE.b (Algorithm \ref{algo:rMLE.b}) &6.04\% (0.12) &60.42\% &3.82\%\\
ONGD (Algorithm \ref{algo:ONGD}) &5.81\% (0.10) &61.94\% &7.52\%\\
\hline
\end{tabular}
\label{tab:CRPS}
\end{center}
\end{table}

The average CRPS obtained by the ideal forecaster over all simulations is 5.78\%. Probabilistic calibration is reflected through a uniform histogram and marginal calibration through the proximity between the predictive and the empirical cdf.s. As expected the forecasts issued by the ideal forecaster are perfectly calibrated in terms of both probabilistic and marginal calibration. On the other hand climatology appears to be significantly not calibrated for all kinds of calibration. Probabilistic persistence performs very well on our synthetic dataset with a CRPS which is already much closer to the CRPS of the ideal forecaster. As one can see in Figure \ref{fig:PIT}, forecasts from probabilistic persistence also essentially achieve probabilistic calibration. However marginal calibration is not quite satisfactory. The last benchmark, that is the rMLE.1 algorithm, performs much better than climatology and better than probabilistic persistence when looking at the CRPS, even with a wrong assumption on the upper bound. However the rMLE.1 forecasts show more departures from probabilistic calibration than probabilistic persistence. 

As for the proposed algorithms \ref{algo:NGD}, \ref{algo:rMLE.b} and \ref{algo:ONGD}, all of them show lower average CRPS.s than probabilistic persistence, the best performance being achieved by NGD and ONGD. Note that ONGD shows a CRPS which is very close to the CRPS of the ideal forecaster, while rMLE.b achieves similar results as rMLE.1 in terms of CRPS. As for probabilistic calibration, ONGD is the algorithm whose histogram is the closest to uniformity, rMLE.b's being clearly not uniform. Only for marginal calibration do all the proposed algorithms show on average better calibration than probabilistic persistence and rMLE.1, especially NGD and ONGD. Overall ONGD is the algorithm which achieves the best results in terms of both sharpness and calibration. In particular it achieves better sharpness and marginal calibration when compared to the already very efficient probabilistic persistence. 

\section{Application to wind power forecasting}
\label{sec:app}
Accurately forecasting wind power generation is highly important for the integration of wind energy into power systems. We are interested here in very short-term forecasting, that is in lead times of a few minutes, which are not only crucial for transmission system operators to keep the system in balance but also very difficult to improve the forecasts for, especially compared to the simple but very efficient persistence.

\subsection{Data description}
We have historical data from a large offshore wind farm, Anholt in Denmark, from July 1, 2013 to August 31, 2014. The active power is available for 110 wind turbines at a temporal resolution of every 10 minute. We scale each time series individually according to the nominal power of the wind turbine and compute the average generation over the wind farm depending on the number of wind turbines which are available at each time step in order to handle missing values. The response random variable we wish to forecast at time $t$ is $X_{t+1} \in (0,1)$, the average active power generated by the wind farm at time $t+1$. As stated in section \ref{sec:sim.forecasting} we choose to look at the observation $x_t$ as a coarsened version of $X_t$ with $\delta=0.001$. Therefore an observation $x_t$ is set to $\delta$ if $x_t < \delta$ and to $1-\delta$ if $x_t > 1-\delta$ and $x_t \in [\delta, 1-\delta]$ whereas $X_t \in (0,1)$. 

\subsection{Validation setup}
\label{sec:val.ANH}
We split our dataset into two datasets that we keep separate: a training/cross-validation dataset from July 1, 2013 to March 31, 2014, resulting in 39,450 observations; a test dataset from April 1 to August 31, 2014, resulting in 22,029 observations. As in section \ref{sec:sim.forecasting} we compare Algorithms  \ref{algo:NGD}, \ref{algo:rMLE.b}, \ref{algo:ONGD} to climatology, probabilistic persistence and to the rMLE.1 algorithm. We use the training set to run the online algorithms, that is the rMLE.1 algorithm and Algorithms \ref{algo:rMLE.b} and \ref{algo:ONGD}, and to train NGD, i.e. Algorithm \ref{algo:NGD}. For methods involving hyperparameters, that is all of them but climatology, we choose the hyperparameters upon cross-validation: we use part of the training set, from November 1, 2013 to March 31, 2014 and for each method select the hyperparameters which give the lowest CRPS on the cross-validation subset. The hyperparameter values selected for each algorithm are available in Table \ref{tab:hyperpar.real}. 
\begin{table}[ht]
\caption{Hyperparameter values for each algorithm: order $p$ of the AR process, forgetting factor $\alpha$, number of iterations $I$, learning rate/step size $\eta$ and minibatch size $m$.}
\begin{center}
\begin{tabular}{lcccc}
&$p$ & $\alpha$ & $\eta$ & $m$\\
\hline
NGD (Algorithm \ref{algo:NGD}) &3 &0.9975 &0.1 &- \\
rMLE.b (Algorithm \ref{algo:rMLE.b}) &5 &0.9982 &- &- \\
ONGD (Algorithm \ref{algo:ONGD}) &4 &- &0.03 &1 \\
\end{tabular}
\label{tab:hyperpar.real}
\end{center}
\end{table}
Note that for Algorithm \ref{algo:NGD} the maximum number of iterations $I$ does not appear in Table \ref{tab:hyperpar.real} as it is not selected upon cross-validation but set to 5,000, which seems to be enough for the objective function to not be significantly decreasing anymore. The frequency in estimating a new model should also be considered as a hyperparameter, as it has an influence on the overall performance of the algorithm. However a serious drawback of Algorithm \ref{algo:NGD} is the associated computing time which is prohibitive. Even if the algorithm runs in a few seconds when $I=5000$, to estimate a new model every 500 data points as in Section \ref{sec:sim} means to estimate 44 models on the cross-validation dataset for one set of hyperparameters. To increase the frequency of the updates for example to an update every 100 data points would mean five times more models for one set of hyperparameters. Therefore we tested only a few update frequencies (100, 250, 500, 750, 1000) and will present the results for 500, which was the frequency that performed the best on the cross-validation set according to this limited grid. This does not mean that a better set of hyperparameters were not to be found if we had infinite computing resources.

\subsection{Assessment of the probabilistic forecasts}
\label{sec:assess.ANH}
\begin{figure}[ht]
    \centering
    \includegraphics[width=\columnwidth]{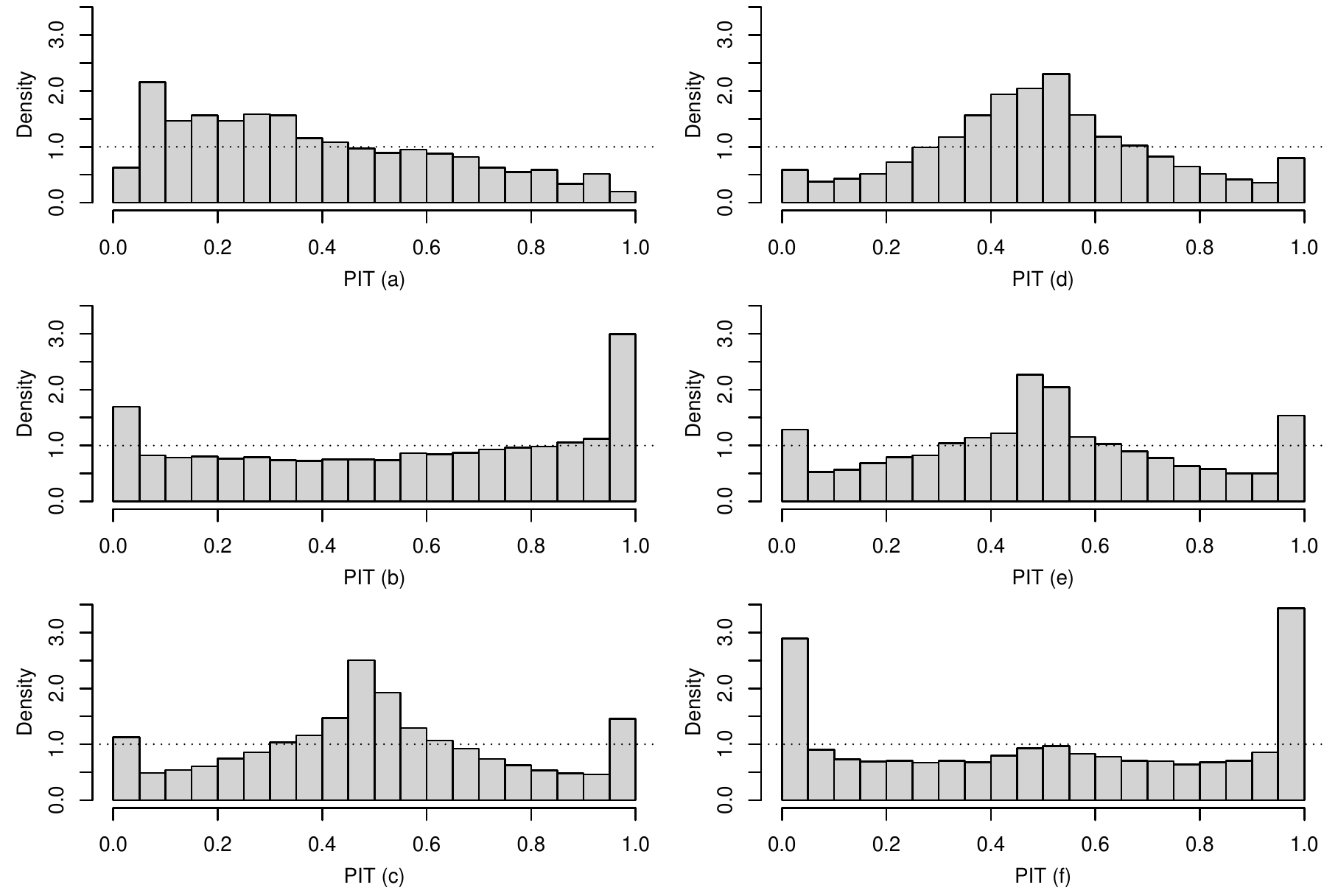}
    \caption{PIT histograms for all benchmarks (left) and Algorithms \ref{algo:NGD}, \ref{algo:rMLE.b}, \ref{algo:ONGD} (right): (a) climatology, (b) probabilistic persistence, (c) rMLE.1, (d) NGD, (e) rMLE.b, (f) ONGD.}
    \label{fig:PIT.ANH}
\end{figure}

\begin{figure}[ht]
    \centering
    \includegraphics[width=\columnwidth]{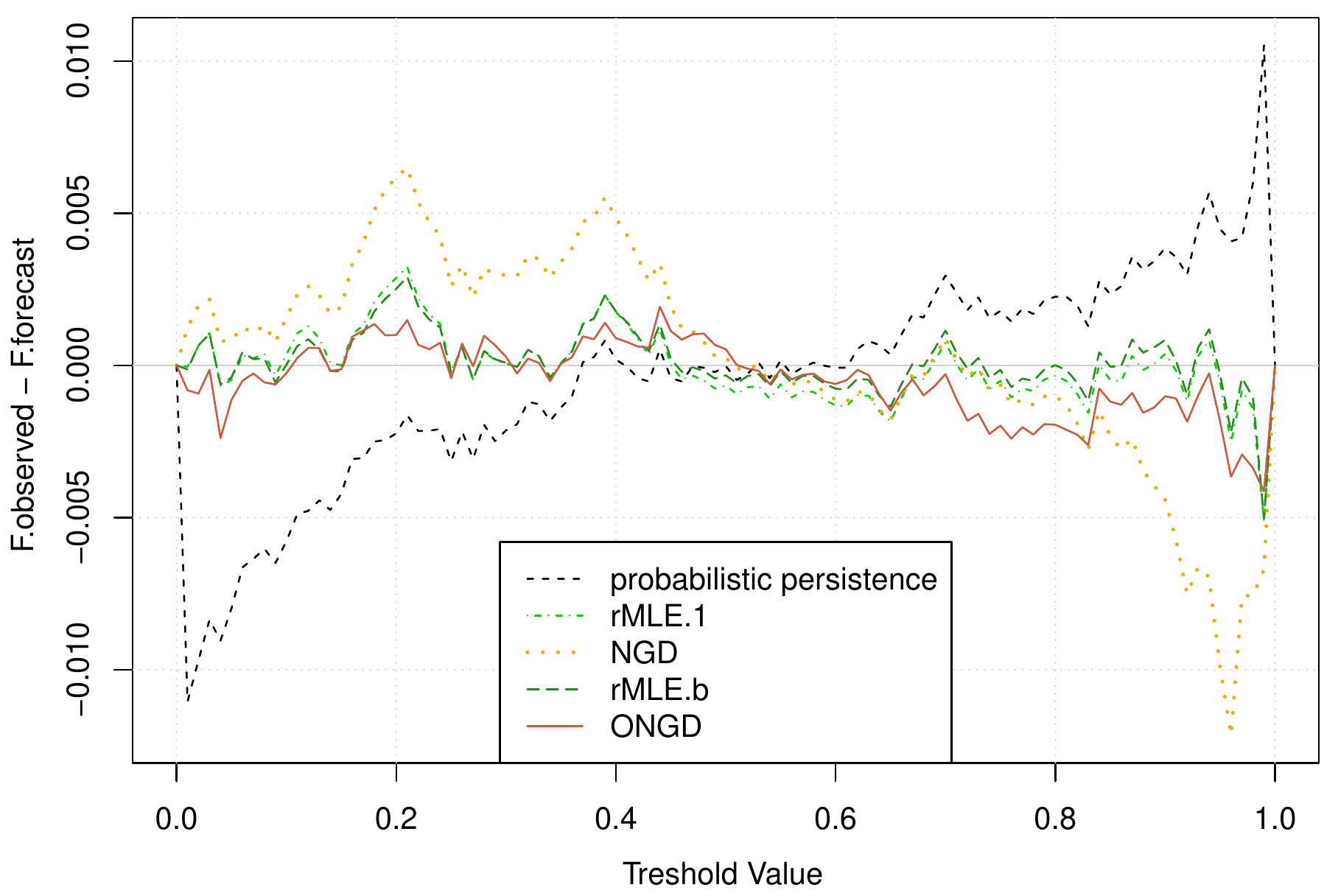}
    \caption{Marginal calibration plot for probabilistic persistence, rMLE.1 and Algorithms \ref{algo:NGD}, \ref{algo:rMLE.b}, \ref{algo:ONGD}.}
    \label{fig:marginal.ANH}
\end{figure}

The algorithms are run on the test set with the hyperparameters we have selected and we evaluate the predictive distributions through calibration and the CRPS as a proper scoring rule, as we did in section \ref{sec:sim.forecasting}. The CRPS are presented in Table \ref{tab:CRPS.ANH} and the calibration plots in Figures \ref{fig:PIT.ANH} and \ref{fig:marginal.ANH}.
\begin{table}[ht]
\caption{10-minute-ahead CRPS and respective improvements over climatology, persistence and the rMLE.1 algorithm.}
\begin{center}
\begin{tabular}{lrrrr}
 &CRPS &Imp./clim. &Imp./persist. &Imp./rMLE.1\\
\hline
climatology &22.03\% &- &- &-\\
probabilistic persistence &1.35\% &93.87\% &- &- \\
rMLE.1 &1.08\% &95.09\% &19.89\% &-\\
\hline
NGD (Algorithm \ref{algo:NGD}) &1.43\% &93.52\% &-5.74\% &-31.99\%\\
rMLE.b (Algorithm \ref{algo:rMLE.b}) &1.06\% &95.21\% &21.83\% &2.43\%\\
\textbf{ONGD} (Algorithm \ref{algo:ONGD}) &\textbf{0.89}\% &\textbf{95.97}\% &\textbf{34.22}\% &\textbf{17.89}\%\\
\hline
\multicolumn{4}{l}{*\textit{Best forecast bolded.}}
\end{tabular}
\label{tab:CRPS.ANH}
\end{center}
\end{table}
On those real data probabilistic persistence improves the CRPS of climatology by a very large percentage already. Nevertheless all algorithms but Algorithm \ref{algo:NGD} perform better than probabilistic persistence in terms of CRPS. One can note that both rMLE algorithms perform quite similarly and improve the persistence by roughly 20\%, while ONGD is the one which achieves the most significant improvement compared to both the persistence and the rMLE algorithms. 
When looking at the calibration plots, it appears that no method is as well calibrated as in the simulation study, no matter which kind of calibration. Regarding probablistic calibration, all PIT histograms suggest departures from uniformity, in a similar way for probabilistic persistence and ONGD, and for rMLE.1 and rMLE.b. The PIT histograms of both persistence and ONGD show a too large number of very low (close to 0) and very high (close to 1) PIT values, which suggests the predictive distributions are underdispersed with too narrow prediction intervals in general. On the contrary the PIT histogram for NGD is hump shaped which indicates the predictive distributions are overdispersed with too large prediction intervals in general. The PIT histograms for the rMLE algorithms somehow show both patterns.
Regarding marginal calibration rMLE.1 and rMLE.b are very close to one another. Overall the online algorithms, that is rMLE.1, rMLE.b and ONGD, show better marginal calibration than the other methods. Note that we even removed climatology from Figure \ref{fig:marginal.ANH} as the corresponding predictive cdf is way too far on average from the empirical one to be plotted on the same graph as the other methods. 

\begin{figure}[!ht]
    \centering
    \includegraphics[width=\columnwidth]{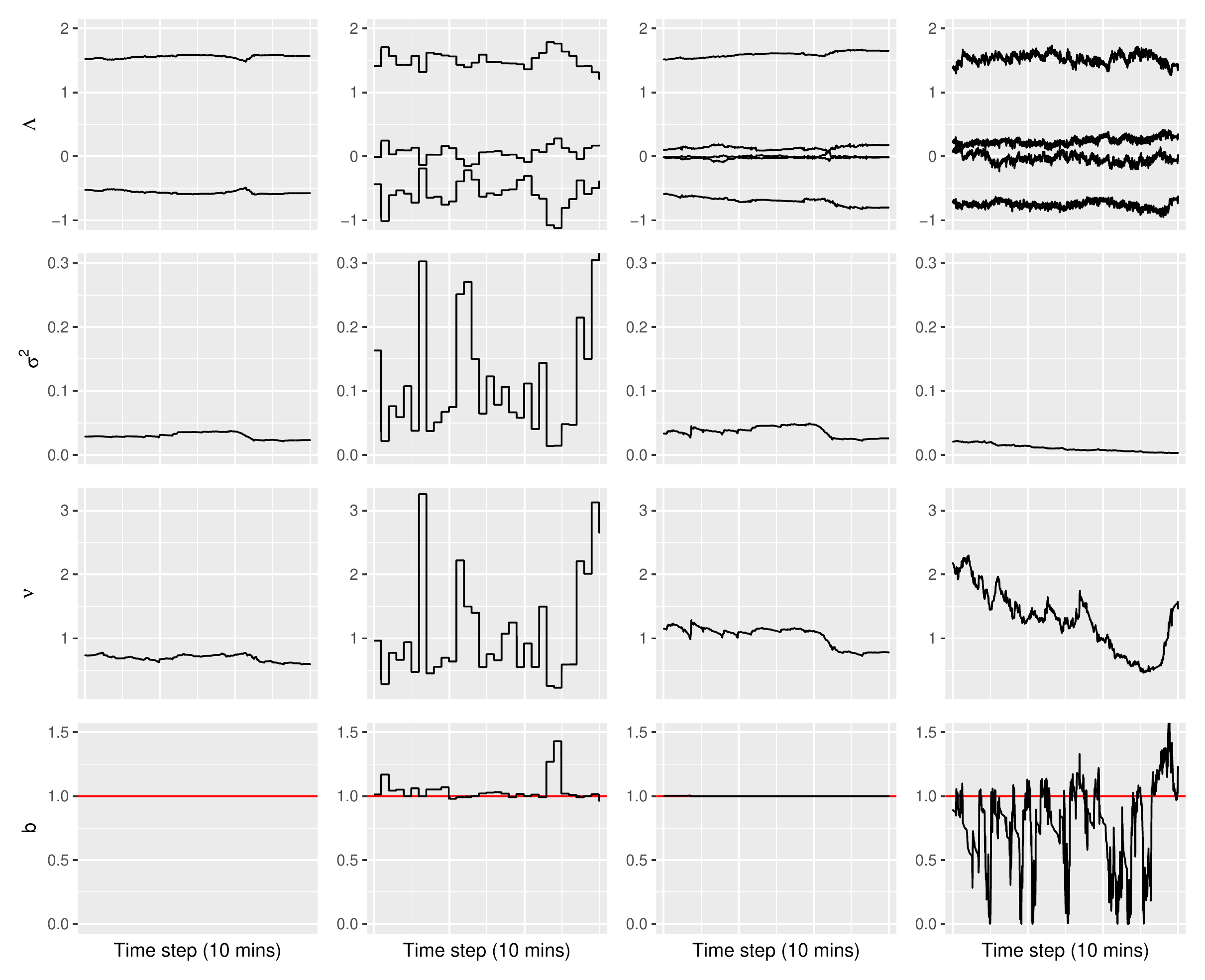}
    \caption{Estimates of $\Lambda$, $\sigma^2$, $\nu$ and projected estimate of $b$ on a sub-sample of the test set for rMLE.1 (left), NGD (center left), rMLE.b (center right) and ONGD (right).}
    \label{fig:par.ANH}
\end{figure}

The parameter vector estimates for the rMLE.1 benchmark and Algorithms \ref{algo:NGD}, \ref{algo:rMLE.b}, \ref{algo:ONGD} are plotted for some sub-sample of the test set in Figure \ref{fig:par.ANH}. Regarding the bound parameter $b$ we plot the projection $\Tilde{b}_t$ of $\Hat{b}_t$, see section \ref{sec:sim.forecasting}, to display the bound which is actually used for prediction. 
One can first note that the estimates of the parameter vector $\Lambda$ are consistent from one method to another, while being more noisy for Algorithms \ref{algo:NGD}, \ref{algo:rMLE.b}, \ref{algo:ONGD} which include a varying upper bound, especially for NGD and ONGD. The estimates from rMLE.1 and rMLE.b are in general very close to one another and show similar patterns with more noise for rMLE.b. Moreover in the latter the estimated upper bound does not vary significantly away from 1. Therefore it is hard to see what rMLE.b brings compared to rMLE.1 on this real dataset. As for Algorithm \ref{algo:NGD}, the CRPS achieved by NGD on the test set was quite high, especially compared to the rMLE.1 benchmark. When looking at its parameter estimates, let first recall that one set of parameters is estimated by one batch model, independently of the other batch models. There is clearly instability from one batch to another, which can be a sign that the maximum number $I=5000$ of iterations we used was not enough for Algorithm \ref{algo:NGD} to converge. Moreover the values estimated for the bound parameter are mostly above 1, which is not satisfactory. Nevertheless the associated computational time would make it very difficult to run the algorithm through even more iterations, as already mentioned in Section \ref{sec:assess.ANH}. Finally ONGD is the only algorithm which captures some variations of the bound parameter below 1 and achieves a very significant improvement over probabilistic persistence (about 32\% reduction of the CRPS) and a significant improvement over rMLE.1 (about 18\% reduction of the CRPS). Although while choosing the hyperparameters for ONGD, we saw a significant improvement in the CRPS on the cross-validation set for a minibatch size $m=1$ only (the values tested for $m$ being $m \in \{1, 5, 10, 20, 50, 100, 150\}$). By looking at the parameter estimates, we noticed that for $m=5$ the bound estimate was also significantly varying below 1, but more slowly and closer to 1, and led to CRPS.s in the range of the other online algorithms. When looking at the power generation itself, the bound tracked by ONGD for $m=1$ did make sense. Those results are confirmed by the test set, as the CRPS we get for it is very close to the one we got on the cross-validation set, confirming the generalization performance of the model. Therefore it seems the data call for a very aggressive choice of $m$ so that the algorithm can track the bound. In return some noise is introduced in the other parameters, especially in the expectation parameter vector $\Lambda$. 
\begin{figure}[!ht]
    \centering
    \includegraphics[width=\columnwidth]{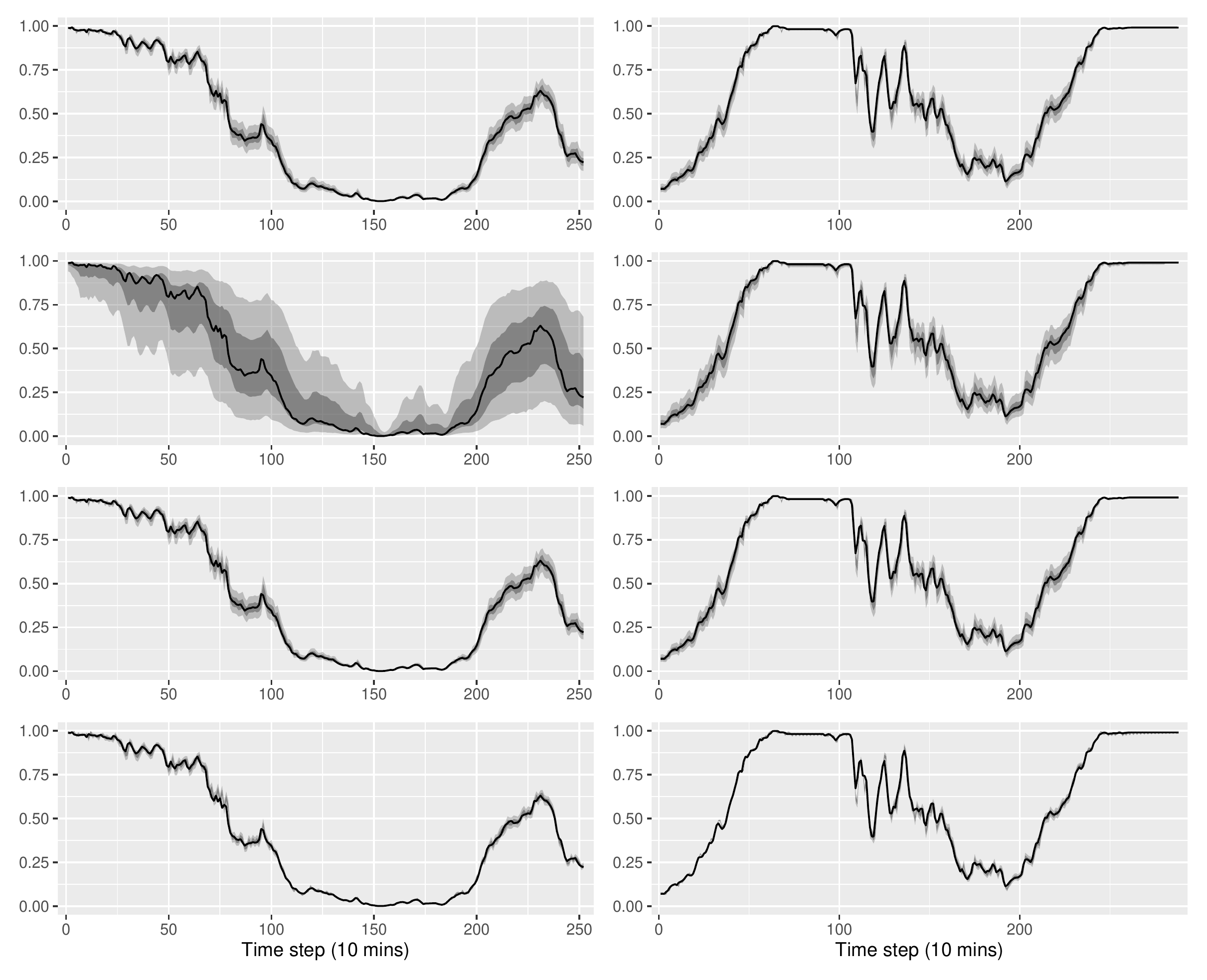}
    \caption{Probabilistic forecasts from rMLE.1 and Algorithms \ref{algo:NGD}, \ref{algo:rMLE.b}, \ref{algo:ONGD} (from top to bottom), on two different periods (left and right), based on prediction intervals with nominal coverage rates of 95 and 75\%, along with the power measurements (solid black line).}
    \label{fig:pred.ANH}
\end{figure}

We provide probabilistic forecasts over two different 36-hour periods of time on the test set for rMLE.1 and Algorithms \ref{algo:NGD}, \ref{algo:rMLE.b}, \ref{algo:ONGD}. As for NGD (second from top) we observe the prediction intervals to be extremely wide on the first period of time (left). This shall correspond to a moment, that is to a batch model, when the scale and/or shape parameters were very large, confirming how problematic the instability from one batch model to another is. As for ONGD (bottom), the plots confirm how tight the prediction intervals are. However when the observation falls out of the prediction interval, which happens quite often as showed by the PIT histogram for ONGD in Figure \ref{fig:PIT.ANH}, it does not fall far away, which explains the very good CRPS achieved by ONGD overall.

\section{Discussion}
\label{sec:discuss}
We have introduced a new framework where we aim to track varying bounds for bounded time series as well as an extended negative log-likelihood to deal with this new framework. As the objective functions now at hand are not convex anymore, we have proposed to make use of the broader quasiconvexity assumption through two algorithms, a batch algorithm and an online algorithm, which both rely on NGD. We have also proposed a more usual online algorithm out of quasi-Newton methods. On both a synthetic and a real dataset, we have run those algorithms for tracking the parameters of a time series distribution over time, including the upper bound of the support of the distribution. Then we have presented how to use those tracked parameters for forecasting. 

The first algorithm, which relies on a time-dependent negative log-likelihood, exponentially weighted through a forgetting factor, and on NGD for its optimization, is a batch algorithm which needs to be updated when new observations come in. It performed well on our (smooth) synthetic dataset but did not scale when moving to our application, that is to wind power forecasting. Extra work could be performed in order to improve it, for example by trying different kinds of initialization for the parameters when starting the optimization of a new batch model. Indeed one could take advantage of the past optimizations by initializing the algorithm with the estimates from the previous batch algorithm. 

The second algorithm is an online algorithm which relies on the same time-dependent, exponentially weighted, negative log-likelihood, but makes use of classical online, local assumptions to recursively update the parameter estimates. It does not perform as well as the first algorithm on our simulated example as it struggles to track an increasing bound and comes with a high price in variance while doing so. It performed well though when tracking a decreasing bound. When moving to wind power forecasting, this second algorithm did not show a significant improvement when compared to its equivalent with a fixed bound. It could be improved by working on adaptive multiple forgetting factors, to adjust to different variation speeds in time and depending on the parameter. This could also benefit to the first algorithm, which performed very well when tracking an increasing bound but was a bit late in following a decreasing one, with some visible impact on both the scale and the shape parameters of the distribution. 

The third algorithm is an online algorithm which is directly derived from SNGD and so, similarly to "ordinary" OGD, only relies on the negative log-likelihood we observe at time $t$ for our current set of parameters. It does not longer require any kind of forgetting action and only asks for the usual step size when updating the parameter vector through the gradient at time $t$. A new hyperparameter which is related to a specificity of SNGD is the size $m$ of the minibatch as SNGD is not guaranteed to converge for $m=1$, unlike SGD. This third algorithm performed extremely well on our simulated examples, with performances in forecasting very close to the ideal forecaster. When moving to wind power forecasting, it improved the CRPS of probabilistic persistence by more than 30\% on the test set. However the predictive distributions do not achieve probabilistic calibration, as the prediction intervals appear to be too narrow in general. 

It is worth noting that ONGD required to set the minibatch size $m$ to 1 on the wind power generation dataset in order to be able to track the bound. This is quite aggressive and suggests that this kind of data might call for methods which can handle big jumps in the bound values. Overall none of the proposed methods nor the benchmarks managed to achieve probabilistic calibration on those data, but looking at the CRPS and at marginal calibration it is quite clear wind power generation forecasting calls for online methods. This is also why we chose not to further investigate NGD on this use case.

\bigskip
\begin{center}
{\large\bf SUPPLEMENTARY MATERIAL}
\end{center}

\begin{description}
\item[Calculation details:] Technical derivations required for Algorithms \ref{algo:NGD}, \ref{algo:rMLE.b}, \ref{algo:ONGD} and detailed computation of the matrix $\Hat{\mathbf{P}}_t$ in Algorithm \ref{algo:rMLE.b}. (.pdf file)

\item[R project:] R project with the R scripts for the simulation study in section \ref{sec:sim}, along with the corresponding synthetic data. All outputs necessary for the study can be reproduced with the corresponding scripts and are also provided. The structure and content of the R project is described in a README file. (.zip file)

\end{description}

\bibliography{mybib}
\end{document}